\documentclass[preprint,12pt]{elsarticle}

\usepackage{amssymb}
\usepackage{amsmath}
\usepackage{url}
\usepackage{booktabs}
\usepackage{multirow}
\usepackage{subfigure}
\usepackage{enumitem}

\setlist[itemize]{noitemsep} 

\usepackage{graphicx} 

\journal{Nuclear Physics B}

\begin{document}

\begin{frontmatter}



\title{Large-scale User Game Lifecycle Representation Learning}


\author[a]{Yanjie Gou} 
\ead{yanjiegou@tencent.com}

\author[b]{Jiangming Liu\corref{cor}}
\ead{jiangmingliu@ynu.edu.cn}

\author[a]{Kouying Xue}
\ead{lornaxue@tencent.com}

\author[a]{Yi Hu\corref{cor}}
\ead{sunracerhu@tencent.com}

\affiliation[a]{organization={Tencent Inc.},
            city={Shenzhen},
            postcode={518000}, 
            state={Guangdong},
            country={China}}

\affiliation[b]{organization={Yunnan University},
            city={Kunming},
            postcode={650504}, 
            state={Yunnan},
            country={China}}      

\cortext[cor]{Corresponding author}
            
\begin{abstract}
The rapid expansion of video game production necessitates the development of effective advertising and recommendation systems for online game platforms. Recommending and advertising games to users hinges on capturing their interest in games. However, existing representation learning methods crafted for handling billions of items in recommendation systems are unsuitable for game advertising and recommendation. This is primarily due to game sparsity, where the mere hundreds of games fall short for large-scale user representation learning, and game imbalance, where user behaviors are overwhelmingly dominated by a handful of popular games.
To address the sparsity issue, we introduce the User Game Lifecycle (UGL), designed to enrich user behaviors in games. Additionally, we propose two innovative strategies aimed at manipulating user behaviors to more effectively extract both short and long-term interests.
To tackle the game imbalance challenge, we present an Inverse Probability Masking strategy for UGL representation learning.
The offline and online experimental results demonstrate that the UGL representations significantly enhance model by achieving a 1.83\% AUC offline increase on average and a 21.67\% CVR online increase on average for game advertising and a 0.5\% AUC offline increase and a 0.82\% ARPU online increase for in-game item recommendation.
\end{abstract}




\end{frontmatter}



\section{Introduction}
The video game industry has undergone rapid growth, now holding a market value in the hundreds of billions of dollars within the entertainment sector.\footnote{\url{https://www.statista.com/topics/868/video-games/.}} This surge necessitates robust game advertising and recommendation strategies for both game industries and online game platforms. The primary goals of these initiatives are to enhance user experiences by advertising games to potential users, recommending in-game items to users, and facilitating suitable user pairings, among other objectives. To realize these goals, it is crucial to analyze user interest in games~\cite{bertens2018machine}.

User interest can be discerned by analyzing their gaming behaviors, such as game logins, ad clicks, and payment tendencies. By summarizing these behaviors, we can derive both sparse and dense features, which form the basis for user interest modeling.
However, according to information theory~\cite{shannon1948mathematical}, the complicated pipeline involved in feature extraction and summarization carries the risk of losing crucial information pivotal for user interest identification. 
Moreover, game advertising and recommendation models are designed for real-world applications, focusing on tasks such as identifying untapped user segments and addressing in-game user needs. These practical applications require feature engineering at both coarse- and fine-grained levels, making the process labor-intensive and time-consuming.

To address the aforementioned challenges and efficiently capture a comprehensive view of user game interest, a natural and promising approach is User Representation Learning~\cite{cen2020controllable,wang2022contrastvae}. This approach directly models user behaviors through self-supervised methods, such as masked item modeling, thereby eliminating the need for complicated feature engineering~\cite{chen2019behavior}.

\begin{figure*}[t]
\centering
\includegraphics[width=1
\columnwidth]{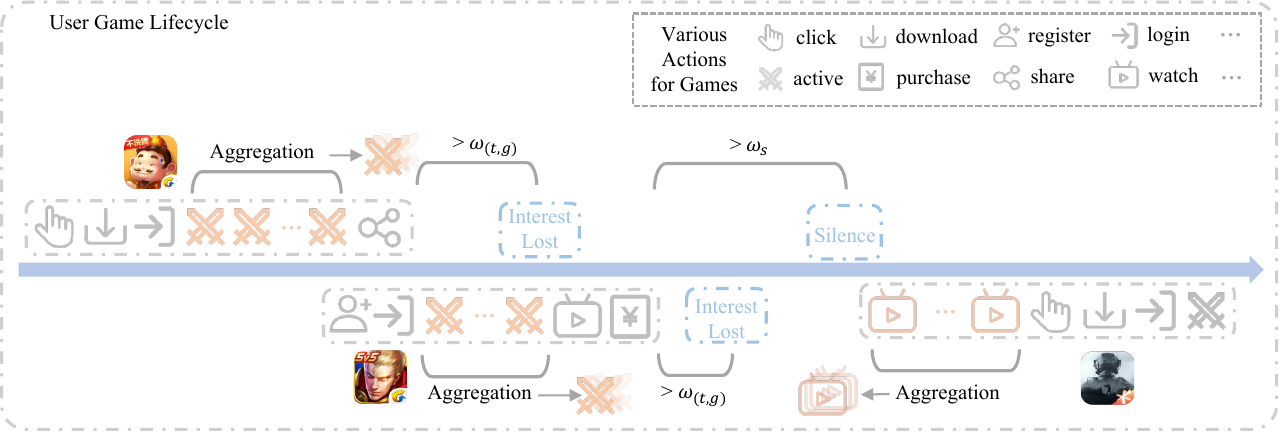}
\caption{An example of User Game Lifecycle with Aggregation and Negative Feedback strategies for user interest induction. $w_{(t,g)}$ is the threshold of the pair of the action type $t$ and the game $g$, and $w_s$ is the threshold for silent actions.}
\label{fig:enhanced-life-circle}
\end{figure*}

However, existing user representation models are primarily designed for item recommendation in scenarios like e-commerce, where users exhibit monotonous behaviors, such as clicking from a vast array of items~\cite{wan2018item,he2018translation}. These models are \textbf{NOT} well-suited for our game scenarios due to two key reasons:
\vspace{0.2cm}
\begin{itemize} 
    \item Game sparsity. Game platforms typically offer a limited selection of games, usually in the hundreds, with each user engaging in only 3-4 of them. This stands in stark contrast to item recommendation scenarios (e.g., e-commerce), where users may interact with thousands of items. Taking these monotonous actions to gaming scenarios is inadequate in accurately capturing user interest.
    \item Game imbalance. User game behaviors exhibit a long-tail distribution. Interactions with popular games overwhelmingly dominate user interest, overshadowing potential interest in other games. This imbalance hinders the effective optimization of user representation.
\end{itemize}
\vspace{0.2cm}
In this paper, we explore effective User Representation Learning in game scenarios, addressing the challenges of game sparsity and game imbalance. To mitigate game sparsity, we introduce User Game Lifecycle (UGL) aimed at enriching user behaviors. As depicted in Figure~\ref{fig:enhanced-life-circle}, UGL enhances user behavior by aggregating diverse game actions from various environments (e.g., platform logins, game or ad clicks, in-game plays, game shares on social media). This approach facilitates a more comprehensive understanding of user interests \textit{both within and outside of games}. These actions are iteratively recorded across multiple games over several days to complete the UGL.
Additionally, we propose two innovative strategies, Aggregation and Negative Feedback, to manipulate user behaviors across different time periods:
\vspace{0.2cm}
\begin{itemize}
    \item Aggregation consolidates recurring actions within a time window to capture consistent interests over a \textit{short} period.
    \item Negative Feedback, a type of non-action feedback, monitors periods of inactivity to explicitly capture a decline in interest over a \textit{long} period.
\end{itemize}
\vspace{0.2cm}
By implementing these strategies, we construct an enriched action sequence for each user, capturing both \textit{short-term} and \textit{long-term} interests across games.

To address the challenge of game imbalance, we propose a self-supervised approach for user behavior modeling called Inverse Probability Masking (IPM). This strategy automatically adjusts the mask probability based on the distribution of user actions across different games, effectively mitigating long-tail bias.
Additionally, by applying representation learning to UGL, the user representation naturally captures user interests at both coarse and fine-grained levels. This unified user representation enhances various downstream tasks, eliminating the need for complicated feature engineering based on expert knowledge of user behaviors.

The contributions of this paper can be summarized as follows:
\vspace{0.2cm}
\begin{itemize}
    \item 
    We explore effective game user representation, addressing the challenges of sparsity and imbalance inherent in the gaming domain.
    \item 
    We introduce UGL, along with the Aggregation and Negative Feedback strategies, to enrich user behavior data and better capture user interests. Additionally, we propose IPM to optimize the representation learning of UGL.
    \item 
    We conduct experiments on game advertising and recommendation tasks, demonstrating the effectiveness of our proposed UGL representation.
    \item 
    Our proposed methods are successfully integrated into Tencent game advertising and recommendation system, yielding significant performance gains.
\end{itemize}
\vspace{0.2cm}

\section{Related Works}
Inspired by recent advancements in self-supervised learning within the Natural Language Processing (NLP) community, the recommendation and advertising industry endeavors to pre-train user behavior models. The goal is to portray both short-term and long-term user interests, thereby enhancing the performance of recommendation and advertising systems. In the subsequent sections, we first introduce related works on user behavior pre-training, then present prevalent pre-training methods in the NLP community, and finally, we introduce the recent advancements in advertising and recommendation systems.

\subsection{User Behavior Pre-training}
For sequential data, foundational self-supervised learning methods are grounded in word2vec~\cite{mikolov2013efficient}. Building on the skip-gram models, \cite{barkan2016item2vec} introduce Item2Vec, which utilizes item-based collaborative filtering to generate item embeddings within a latent space. In Item2Vec, a user representation can be derived by aggregating the interacted items of a given user. Based on Item2Vec and under the assumption that user behaviors and interests are influenced by the items presented to the user, \cite{barkan2020attentive} apply the neural attention mechanism to enhance the capabilities of Item2Vec.

A self-supervised user modeling network is proposed to encode user behavior data into universal representations by integrating a behavior consistency loss. This loss guides the model to fully identify and preserve valuable user information~\cite{gu2021exploiting}. Building upon this work, additional self-supervised tasks of masked behavior prediction and next $K$ behavior prediction are introduced to aid in model training~\cite{wu2020ptum}. The former can model the relatedness between historical behavior, while the latter can model the relatedness between past and future behaviors.

Taking a different approach, \cite{qiu2021u} incorporate a review encoder to model user behaviors, alongside a user encoder integrating product reviews. On the other hand, \cite{wu2022userbert} introduce UserBERT to learn universal user models on unlabeled user behavior data. This is accomplished through two contrastive self-supervision tasks: masked behavior prediction and discrimination. These tasks focus on modeling the contexts of user behaviors to capture user interest across different periods. 
Additionally, \cite{cheng2021learning,xie2022contrastive} introduce contrastive learning to user behavior modeling. \cite{bian2021contrastive} combine contrastive learning with curriculum learning for producing effective representations for modeling sequential user behaviors via data quality and sample ordering. \cite{wang2023missrec} propose a multi-modal pre-training and transfer learning framework to capture the sequence level multi-modal user interest.

However, these methods fall short in achieving satisfactory performance in game recommendation and advertising, and the reasons for this are twofold.
Firstly, unlike in recommendation systems where each user can interact with thousands of items, in the gaming context, users typically play only 3-4 games on average. When we apply actions commonly used in recommendation systems (e.g., clicks and purchases) to model user interest in games, the resulting action sequences become monotonous and repetitive.
Secondly, the frequency of actions related to games suffers from a severe long-tail issue, wherein actions associated with popular games overwhelmingly dominate user behavior. 

Some existing works focus on addressing the long-tail problem in recommendation systems~\cite{bai2017dltsr,liu2020long}, targeting either long-tail user \textbf{OR} item. \cite{yin2020learning} tackle the long-tail user problem by learning transferable parameters from both optimization and feature perspectives with a gradient alignment optimizer and adopt an adversarial training scheme. \cite{zhang2021model} propose a dual transfer learning framework to mitigate the long-tail item problem by jointly learning the knowledge transfer from both model-level and item-level. In contrast, the game scenarios face with \textbf{BOTH} the long-tail user and item problems, specifically game sparsity and imbalance. To this end, we propose data enhancement strategies and Inverse Probability Masking method.

\subsection{Pre-training Paradigms}
The emergence of self-supervised pre-training marks a new era for the NLP community. Built upon the Transformer architecture, pre-training paradigms can be categorized into three distinct types: encoder-only, decoder-only, and encoder-decoder pre-training. Leveraging the Transformer encoder, a prevalent self-supervised task is masked language modeling~\cite{devlin2018bert,liu2019roberta}, which has been adoptted in the recommendation and advertising industry for modeling user sequence behavior~\cite{wu2020ptum,wu2022userbert}.

Based on the Transformer decoder, a common self-supervised task involves predicting the next token~\cite{radford2018improving,radford2019language}. This task has also been leveraged by some works in recommendation systems~\cite{sileo2022zero,cui2022m6}.
Taking both the Transformer encoder and decoder, \cite{lewis2020bart} introduce five methods (e.g., sentence permutation and token deletion) to corrupt the original text. The model then recovers the lost information through a sequence-to-sequence pre-training method. Additionally, \cite{raffel2020exploring} propose T5, which utilizes the BERT-style masking strategy in spans. Inspired by T5, \cite{geng2022recommendation} introduce P5 that treats recommendation as language modeling within a unified pre-training, personalized prompting, and prediction paradigm.

Our method is inspired by masked language modeling~\cite{devlin2018bert}, 
which has demonstrated effectiveness in extracting information from billion-level sequence data. We extend this approach to game advertising and recommendation systems through large-scale representation learning.

\subsection{Advertising and Recommendation}
Due to the richness of scenarios in advertising and recommendation systems, different contexts exhibit distinct characteristics. Consequently, numerous methods have emerged to address the requirements specific to each scenario.
E-commerce is the most prevalent application scenario for advertising and recommendation systems, attracting significant attention within the recommendation research community. For instance,
\cite{wang2023sequential} equip sequential item recommendation with a novel causal discovery module to capture causalities among user behaviors. 
\cite{wu2023instant} propose a graph neural network for dynamic multiplex heterogeneous graphs, which develops a sample-update-propagate architecture to alleviate neighborhood disturbance.

Also, there are other platforms that need effective recommendation systems. 
For example, in the online group buying application, the correlations and interactions between initiators, participants and items are complicated, \cite{zhai2023group} address this by introducing multi-task learning.
For live channel recommendation, the scenario faces issues such as complex and inconsistent user intentions across different domains,
\cite{zhang2023cross} employ the disentangled encoder module to learn user’s cross-domain consistent intentions and domain-specific intentions.
For online travel platforms, flight recommendation faces the challenge of exploring origin and destination cities as well as learning about them as a whole. \cite{xu2022odnet} propose a personalized origin-destination ranking network to address this issue.
For news recommendation, dynamic nature of a news domain causes the
problem of information overload that makes it difficult for a
user to find her preferable news articles. To address this issue,
\cite{lim2022airs} introduce how the NAVER News service conduct new recommendation, where collaborative filtering, quality estimation, social impact models are involved.

In contrast to the aforementioned scenarios, game advertisement and recommendation faces its unique challenges, i.e., game sparsity and game imbalance. These challenges stem from the inherent nature of the gaming domain, where the distribution of user preferences and interactions can be highly skewed and sparse. As a result, it is essential to develop tailored solutions that specifically address the sparsity and imbalance issues in the context of game advertisement recommendation.

\section{User Game Lifecycle} \label{sec:3}
A large-scale user base with sparse and imbalanced individual behaviors leads to data sparsity in game advertising and recommendation scenarios. Employing effective data augmentation methods to generate meaningful user representations is crucial. In this section, we first present several definitions and then describe the proposed solution, i.e., UGL, which is used to collect and enrich user behaviors.

\subsection{Definitions} \label{subsec::act} 

\vspace{0.2cm}
\noindent
\textbf{Definition 1} (Basic Action). \textit{A basic action $a$ is a triple $(t, g, d)$, where $t \in T$ is an action type, $g \in G$ is the game on which the action is taken, and $d \in D$ is the date when the action is taken.}

Based on the action environments, action types are categorized into in-game actions (e.g., playing games and paying for virtual items) and actions taken outside of games (e.g., clicking on game ads), as they strongly indicate potential interest in games. 
We gather actions from four categories: In-Game Activity ($T_{iga}$), Search Scenarios ($T_{ss}$), Advertisement and Recommendation ($T_{ar}$), and Social Media and Community ($T_{smc}$), where $T = T_{iga} \cup T_{ss} \cup T_{ar} \cup T_{smc}$.

\subsubsection{In-Game Activity} Game playing is a primary action that directly reflects the gaming intention of users, providing direct and real-time insights into their preferences, behaviors, and engagement patterns. Users can click on game apps, log in to games, and make in-game purchases. By analyzing in-game activity, game recommendation systems can gain a nuanced understanding of user interest in gaming. So we collect all these types of behaviors as user in-game activity actions ($T_{iga}$). Additionally, game-specific actions, such as participating in competitions in Multiplayer Online Battle Arena (MOBA) games, are also included.

\subsubsection{Search Scenarios} Beyond game playing, users usually search for information about the games they have played (e.g., game walkthroughs and puzzle solutions). They take actions like clicking on the content provided by search engines and consuming content (e.g., articles, videos, and accessories). These actions are \textit{intrinsically motivated}. We collect all these behaviors as search scenearios actions $T_{ss}$.

\subsubsection{Advertisement and Recommendation.} Game advertisement and recommendation aim to promote games to potential users and display related items to target users, where reactions to them are \textit{extrinsically motivated}. In game advertisements, users can click on ads, download game apps, make appointments for game releases, and register for games. In game recommendations, users can click on items, make purchases, etc. We collect all these behaviors as advertisement and recommendation actions $T_{ar}$.

\subsubsection{Social Media and Community.} Social media and community platforms (e.g., Imagine Games Network) are popular venues for users to discuss games. For example, they can update their status about game playing and share games with their friends; they explicitly express their likes or dislikes on the game reviews and comments. These behaviors are essential for establishing \textit{social connections} among users. We collect these behaviors as social media and community actions $T_{smc}$.

\vspace{0.2cm}
\noindent
\textbf{Definition 2} (Negative Action). \textit{A negative action $na = (t, g, d)$ is generated by monitoring the inactivity of
basic actions in a period of time. The negative action can be a lost action $la$ or a silence action $sa$.}

The negative feedback strategy is proposed by incorporating a form of non-action feedback that captures the loss of user interest over a period of time by monitoring the duration of inactivity. This approach enables the detection of changes in user interest. This operation consists of two types of feedback: \textit{lost} and \textit{silence} feedback.

\vspace{0.2cm}
\noindent
\textbf{Definition 3} (Lost Action). \textit{A lost action $la = (t, g, d)$ is a specific type of negative action. A lost action is generated if the time gap between the same continued basic actions is larger than a threshold. 
Formally, the basic action sequence is $[(t_i, g_i, d_i), \ldots, (t_j, g_j, d_j)]$ ($i<j$), and if $(t_i, g_i) = (t_j, g_j)$ and $(t_i, g_i) \neq (t_k, g_k)$ for $(i<k<j)$ and $d_j - d_i > \omega_{(t_i,g_i)}$, where $\omega_{(t_i,g_i)}$ is a lost threshold, the original sequence will be enriched as $[(t_i, g_i, d_i), \underline{(\bar{t}_i, g_i, d_i)}, \ldots, \\ \underline{(\bar{t}_{j}, g_{j}, d_{j})}, (t_{j}, g_{j}, d_{j})]$, where the underlined ones are the lost actions.}

In a gaming scenario, if a user takes an action on a game by mistake, the system will have the action recorded, although the action may not accurately represent the user interest. The lost action can provide negative feedback to capture such mistakes. Additionally, a period of time gap between basic actions is modeled as patterns of user actions.

\vspace{0.2cm}
\noindent
\textbf{Definition 4} (Silence Action). \textit{A silence action $sa = (t, g, d)$ is a specific type of negative action. A silence action is generated if the time gap between neighboring basic actions is larger than a threshold. For the basic action sequence $[(t_i, g_i, d_i), (t_j, g_j, d_j), \ldots]$ ($i<j$), if $d_j - d_i > \omega_{s}$, where $\omega_{s}$ is a silence threshold, the original sequence will be enriched as $[(t_i, g_i, d_i), \underline{(o, o, d_i)}, \underline{(o, o, d_j)}, \\ (t_j, g_j, d_j), \ldots]$, where $o$ is a special symbol representing the silence action type and corresponding games, and the underlined tuples signify that a period of silence occurred between the dates $d_i$ and $d_j$.} \footnote{The special symbol $o$ is added to the set of action types $T$ and the set of games $G$.}

Unlike the lost action that is generated at the game level, the silence action is generated at the action-type level for all games. If a user does not take any action for all games, the silence action will be generated, indicating that the user has no interest in any game for a period of time. These patterns intuitively contribute to capturing the long-term user interest.


As discussed, the determination of these thresholds depends on the specific definitions of ``lost" and ``silent" within the given context. In gaming, a user who does not interact with a game for 7 days is typically considered to have lost interest. Incorporating this ``lost'' action into the user behavior sequence helps enhance user representation by accounting for disengagement. The same principle applies to the ``silent" action.
In this work, the thresholds $\omega_{(t_i,g_i)}$ and $\omega_{s}$ are both set to 7 days, which are determined through a grid search across various thresholds (1, 3, 7, 14, and 30 days). The 7-day threshold is found to be optimal, outperforming the other thresholds. This suggests that a 7-day threshold effectively balances sensitivity to user inactivity with the stability of user behavior patterns, aligning with our initial insights.


\vspace{0.2cm}
\noindent
\textbf{Definition 5} (Aggregated Action). 
\textit{An aggregated action $x$ is extended by a basic action or a negative action $x = (t, g, s, e, f)$, where $s$ is the start date of the action, $e$ is the end date of the action, and $f$ is the frequency of the action type $t$ on game $g$ in the period $[s, e]$.}

We propose the aggregation strategy to generate aggregated actions based on corresponding basic actions, aiming to further refine user interest. As illustrated in Figure ~\ref{fig:enhanced-life-circle}, actions that are repeatedly taken on the same game within a time window represent \textit{short-term} interest. Specifically, an aggregated action $(t, g, s, e, f)$ is induced if and only if the basic action type $t$ on game $g$ occurs exactly $f$ times within a period of $[s, e]$, and no other actions are taken during that period.

Aggregation primarily focuses on aggregating repetitive action types such as game login and play, which occur frequently and can overwhelm other actions. Therefore, Aggregation can derive concise representations of short-term interest and eliminate redundant information, thereby improving data quality.

\vspace{0.2cm}
\noindent
\textbf{Definition 6} (User Game Lifecycle).
\textit{A User Game Lifecycle (UGL) is represented as a sequence of user aggregated actions in various game environments over time, aiming to increase the diversity of user behaviors, denoted as $O = [x_0, x_1, x_2, \ldots, x_{N-1}]$, where $x_i$ is the $i$th aggregated action, and $N$ is the number of actions.}

Due to the complexity and heterogeneity of user behaviors on games, the User Game Lifecycle unifies these behaviors using predefined actions. The actions are collected from multiple environments to provide a comprehensive understanding of user intentions at both coarse- and fine-grained levels, encompassing diverse contexts within and outside of games.



\begin{figure}[t] 
\centering
\includegraphics[width=0.6\columnwidth]{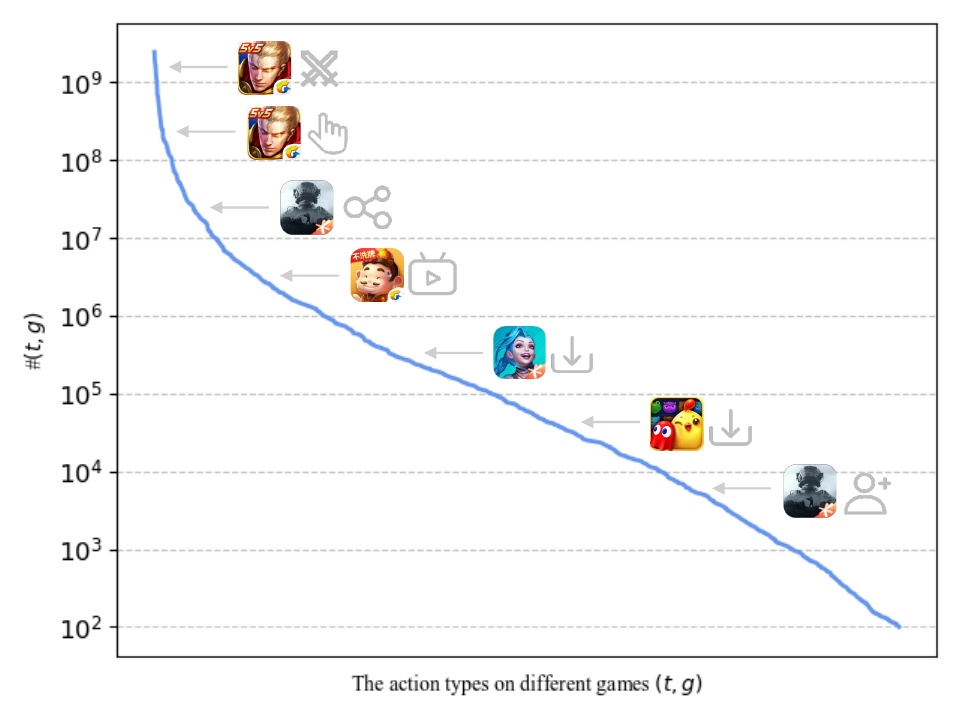}
\caption{
The distribution of action types across different games, where $(t,g)$ represents the action with type $t$ in the context of game $g$. Notably, actions associated with popular games (on the left side) exhibit a significant prevalence within the sequence data, while actions related to cold games (on the right side) are relatively sparse.}
\label{fig:longtail}
\end{figure}

\subsection{Long-tailed Phenomena in Game Actions}
The long-tailed phenomena in game actions are caused by the game imbalance issue, where a small portion of $(t, g)$ highly frequently occurs, dominating the distribution over action types and games. This phenomenon is referred to as long-tailed phenomena in game actions.

As shown in Figure~\ref{fig:longtail}, the number of $(t, g)$ on the left-hand side of the x-axis is significantly more frequent than that on the right-hand side of the x-axis. The $(t, g)$ on the left-hand side dominate the sequence patterns, making it challenging to distinguish users based on the $(t, g)$ on the right-hand side. The $(t, g)$ on the left-hand side occur in popular games, while the $(t, g)$ on the right-hand side occur in less popular games.

However, recent works adopt the standard masking strategy~\cite{wu2020ptum,wu2022userbert}, which applies the same probability to mask tokens without distinguishing $(t, g)$ in popular and less popular games. This approach results in a bias that overfits to popular games but underfits to less popular games, making it difficult to effectively model user interest displayed by actions on less popular games. 

To tackle this issue, we propose Inverse Probability Masking (details are shown in Section~\ref{subsec:4.2}), which balances the masking probability across actions on games. With this strategy, actions in both popular and less popular games are effectively exploited during representation learning, achieving a balanced optimization over the representation space and mitigating the long-tail bias.


\begin{figure}[tb]
  \centering
    \includegraphics[width=\columnwidth]{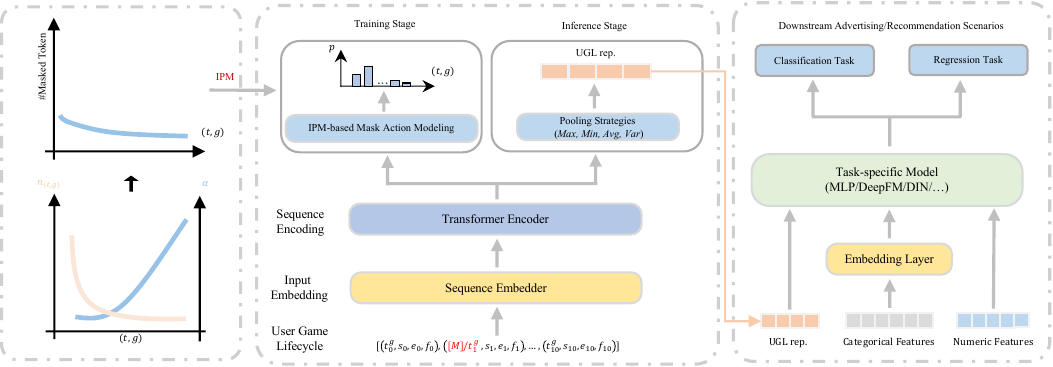}
  \caption{Illustration of Inverse Probability Masking (IPM), a model structure, and the application of UGL rep.\ in a task-specific model. In the model structure, $t^g_i$ is masked as [\textit{M}] during the training stage, and $t^g_i$ remains unmasked during the inference stage. Notably, in the training stage, IPM is applied to optimize UGL representation learning. The UGL rep.\ obtained in the inference stage is fed to a task-specific model for a downstream advertising/recommendation task.}
\label{fig:encoder}
\end{figure}

\section{Model}
In this section, we first present the model structures for self-supervised UGL representation learning and then introduce masked action modeling with the proposed Inverse Probability Masking (IPM). Based on the fully-trained UGL models, we infer the user UGL representations that are used for advertising and recommendation tasks.

\subsection{Model Structure} 
The model structure consists of a \textit{sequence embedder} and a \textit{sequence encoder}, which is shown in the middle of Figure~\ref{fig:encoder}. A sequence embedder first obtains the representation of aggregated action attributes (i.e., action type, game, start date, end date, and frequency) and then collects them to be input embeddings. Taking the input embeddings as input, a sequence encoder adopts Transformer to obtain contextual representations given the input embeddings. The details of the two components are presented below.

\subsubsection{Sequence Embedder} 
\begin{figure*}[t]
\centering
\includegraphics[width=\columnwidth]{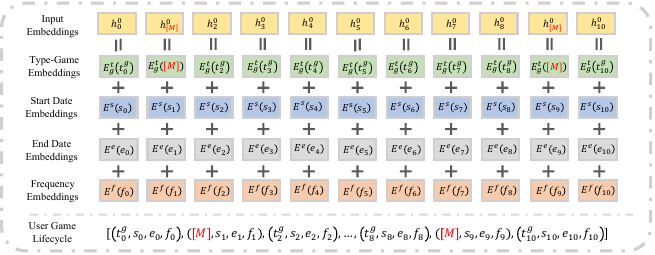}
\caption{
Illustration of the Sequence Embedder. The input embedding is computed as the sum of embeddings for type-game, start and end dates, and frequency.}
\label{fig:embedder}
\end{figure*}

As shown in Figure~\ref{fig:embedder}, 
the discrete actions in UGL are encoded into dense vectors by a sequence embedder. Given a UGL $O = [x_0, x_1, x_2, \ldots, x_{N-1}]$ (a sequence of aggregated actions), according to five attributes of aggregated actions, five separate sequences are generated: a type sequence $[t_0, t_1, t_2, \ldots, t_{N-1}]$, a game sequence $[g_0, g_1, g_2, \ldots, g_{N-1}]$, a start date sequence $[s_0, s_1, s_2, \ldots, s_{N-1}]$, an end date sequence $[e_0, e_1, e_2, \ldots, e_{N-1}]$, and a frequency sequence $[f_0, f_1, f_2, \\ \ldots, f_{N-1}]$. For simplification, we combine the type sequence $[t_0, t_1, t_2, \ldots, t_{N-1}]$ and the game sequence $[g_0, g_1, g_2, \ldots, a_{N-1}]$ into a single sequence for action types and games $[t^g_0, t^g_1, t^g_2, \ldots, t^g_{N-1}]$, where $t^g_i$ denotes the action type $t_i$ taken on game $g_i$. Finally, we obtain four kinds of sequences for a UGL. 

For these sequences, we adopt lookup tables to build the type-game embedder $E^{t}_g(\cdot)$, start date embedder $E^s(\cdot)$, end date embedder $E^e(\cdot)$, and frequency embedder $E^f(\cdot)$, to obtain corresponding embeddings, respectively. The embeddings of the user action sequence, $H^0 = [H^0_0, H^0_1, H^0_2, \ldots, H^0_{N-1}]$, are computed as follows:
\begin{equation}
H^0_i = E^t_g(t^g_i) + E^s(s_i) + E^e(e_i) + E^f(f_i),
\end{equation}
where $H^0_i$ is the embedding of the $i$th action.

\subsubsection{Sequence Encoder} 
The embeddings of the user action sequence, $H^0$, serve as the input to a sequence encoder. The encoder outputs contextualized representations using the self-attention mechanism and feed-forward networks:
\begin{equation}
H^l = \textsc{TransformerLayer}_l(H^{l-1}), \quad l \in [1, L],
\end{equation}
where $\textsc{TransformerLayer}_l(\cdot)$ is the $l$-th layer of the Transformer, $L$ is the number of stacked layers, and $H^L$ is the representation from the $L$-th layer of the Transformer. 

$H^L = [h^L_0, h^L_1, h^L_2, \ldots, ...h^L_{N-1}]$ is rooted from the corrupted action sequence and is used for model training with mask action modeling (Section~\ref{subsec:4.2}). On the other hand, the uncorrupted action sequence will be used when aiming to obtain the user representation in the inference stage with pooling strategies (Section~\ref{subsec:4.3}).

\subsection{Training}
\label{subsec:4.2}
Given the training data consisting of users paired with their historical behaviors across various game environments, we collect and construct a User Game Lifecycle (UGL) for each user according to the definitions given in Section~\ref{sec:3}.

We propose masked action modeling for self-supervised learning to train the UGL models. Given a type-game sequence $[t^g_0, t^g_1, t^g_2, ..., t^g_{N-1}]$, we mask items with probability $q$. The masked action types and games are replaced with the special symbol [\textit{M}], and simultaneously recovered using the contextual information of unmasked tokens.

Formally, the masked token is encoded as $h^L_{[M]}$. It is then passed to the prediction layer as follows:
\begin{equation}
p([M]) = \textsc{SoftMax}(W^T h^L_{[M]} + b),
\end{equation}
where $p([M]) \in \mathbb{R}^{|T|\cdot|G|}$ is the posterior distribution for the masked token, and $|T|$ and $|G|$ are the number of action types and games, respectively. $W$ and $b$ are the learnable parameters. The UGL model is optimized by minimizing the cross-entropy loss function:
\begin{equation}
L = -\sum \log p([M] = t^g_{[M]}).
\end{equation}

\subsubsection{Inverse Probability Masking}\label{subsec:ipw}
Instead of the constant masking probability $q$, Inverse Probability Masking (IPM) assigns each pair of action type and game with a particular masking probability according to the distribution of action types and games, so as to address the imbalance in different actions.
In other words, IPM effectively smooths out extreme distribution differences by mitigating the impact of high-frequency game actions and amplifying the significance of low-frequency game actions, resulting in more robust user representations. 
By doing so, it ensures that less frequent actions receive more attention during the training process, leading to a more balanced and representative model.

The masking probability $q^{\text{IPM}}_{(t,g)}$ for the action type and game $(t, g)$ is computed as follows:
\begin{align}
    q^{\text{IPM}}_{(t, g)} =& \left\{
    \begin{aligned}
    &q^c & \text{if}\  \alpha_{(t,g)} q^v >= q^c \\
    &\alpha_{(t,g)}  q^v & \text{if}\  q^c > \alpha_{(t,g)}  q^v > 0 \\
    \end{aligned}
    \right. ,
\end{align}
where, $q^c$ and $q^v$ are constant probabilities, and $\alpha_{(t,g)}$ is a scaling coefficient used to control the masking probability for the pair of action type and game $(t,g)$. $\alpha_{(t,g)}$ is computed as follows:
\begin{equation}
    \alpha_{(t,g)} = \frac{1}{m_{(t,g)}  |T|  |G|},
\end{equation}
where $m_{(t,g)}$ is the proportion of $(t,g)$ and is computed as follows:
\begin{equation}
    m_{(t,g)} = \frac{n_{(t,g)}}{\sum_{a'}{\sum_{g'}{n_{(t',g')}}}},
\end{equation}
where $n_{(t,g)}$ is the number of $(t,g)$ in the training data.
As shown in the left bottom of Figure~\ref{fig:encoder}, a smaller $m_{(t,g)}$ leads to a higher $\alpha_{(t,g)}$, resulting in a higher masking probability $q^{\text{IPM}}_{(t,g)}$. Additionally, a constant probability $q^c$ is used to bound the masking probability to a reasonable value. This ensures that the number of masked tokens is similar for all pairs of action types and games, as shown in the left top of Figure~\ref{fig:encoder}. 
The key insight of IPM is that the importance of actions often correlates with their rarity. This means that the most valuable actions are typically the least frequent. For example, in Conversion Rate (CVR) scenarios, actions such as clicks, downloads, and registrations for an item usually decrease in frequency in that order.


\subsection{Inference} \label{subsec:4.3}
Intuitively, employing various pooling techniques entails extracting the information embedded within $H^L$ from distinct perspectives. Therefore, we employ four pooling strategies: max pooling ($\textsc{Max}(\cdot)$), average pooling ($\textsc{Avg}(\cdot)$), minimization pooling ($\textsc{Min}(\cdot)$), and variance pooling ($\textsc{Var}(\cdot)$) to provide a comprehensive understanding of the user representation. Variance pooling executes the variance operation dimension-wise, serving as a measure of the degree of change in the action sequence. 

The final UGL representation is obtained by concatenating the outcomes of the aforementioned pooling techniques:
\begin{equation}
u = [\textsc{Max}(H^L); \textsc{Avg}(H^L); \textsc{Min}(H^L); \textsc{Var}(H^L)]. 
\end{equation}
Beyond the utilized pooling strategies in this paper, more advanced techniques for effectively aggregating valuable information from $H^L$ are left for future works.

By conducting inference on all available game users in the dataset based on the UGL, we construct an offline database of game user representations, i.e., \(U = (u_1, u_2, \ldots, u_i, \ldots)\).

\subsection{Task-Specific Models}
The task-specific models aims to classify if the given user is a targeted user for advertising and recommendation. As shown in the right of Figure~\ref{fig:encoder}, the task-specific models take the user features as the input, including the categorical features (e.g., occupation and gender) and the numeric features (e.g., age and payment). Additionally, the UGL representations are taken into the models as extra features.

\begin{table}[t]
  \centering
    \begin{tabular}{lrrrr}
    \toprule
    Dataset & \#User & \#Item & \#Sparse & \#Dense \\
    \midrule
    UGL Pre. & 6 million & 24,851 & / & /  \\
    Game Ad. & 8 million & 20 & 283 & 364  \\
    Game Rec. & 0.4 million & 259 & 454 & 317  \\
    \bottomrule
  \end{tabular}
  \caption{The dataset information of UGL Pre-training (UGL Pre.), Game Advertising (Game Ad.) and In-Game Item Recommendation (Game Rec.).}
  \label{tab:data_info}
\end{table}

\section{Experiments}
We conduct offline and online experiments to validate the effectiveness of our UGL representations. Our experiments focus on game advertising and in-game item recommendation tasks to demonstrate that our proposed method can automatically learn user interests that are helpful for building game advertising and in-game item recommendation system. Additionally, we compare our user representation learning approach with other existing methods to highlight the superiority of our methods in the game scenarios.

\subsection{Settings for UGL Representation Learning}
\subsubsection{Data}
In the task of user representation learning, we utilize real industrial data to perform User Game Lifecycle (UGL) for all users. The UGL is constructed by arranging the aggregated actions in chronological order, from the earliest to the most recent actions, encompassing various action types across 300 games from July 2022 to January 2023.
We incorporate more than 50 action types.
The maximum length of the UGL sequence is set to 128. Truncation is applied if the sequence length exceeds 128, while padding is applied if the sequence is shorter than 128. In our experiments, we collect data from 600 million users using real industrial data and obtain an offline database of UGL representations for these users.


\subsubsection{Model and Training} 
In our user representation learning task, we set the dimension of both input embedding and hidden representation to 64. The number of Transformer encoder layers is 6, and the dimension of the UGL representation is 256. For optimization, we use the Lamb optimizer with an initial learning rate of \(1.76 \times 10^{-3}\). The training step is set to 300,000 with a batch size of 4,096. We apply \(l2\)-regularization with a weight of 0.01. In the Inverse Probability Masking (IPM), the constant probabilities \(q^c\) and \(q^v\) are set to 0.15 and 0.5, respectively. The UGL models are trained on 4 NVIDIA V100 32GB GPUs for a duration of 200 hours.

\subsubsection{Use in Downstream Tasks}
The fully-trained UGL representations are integrated into the downstream models as additional features. The downstream tasks involve searching for the UGL representation based on unique user IDs. If a user cannot be found in the set of UGL representations, we use default values of 0s.

\subsection{Settings for Game Advertising}
In the game advertising task, our goal is a classification task to determine whether users will download and register the game displayed to them. The classification of games into popular, moderate, and cold categories is based on the number of monthly active users. A game is considered popular if it has a large-scale user base, i.e., more than 10 million monthly active users; it is classified as moderate if there are around 1 million monthly active users, and it is considered cold if it has less than 1 million monthly active users. The scale of the game influences the frequency of game-related behaviors.
We use real-world video games in our experiments, specifically game \textit{A} and \textit{B} as popular games, \textit{C}, \textit{D}, \textit{E} as moderate games, and \textit{F} and \textit{G} as cold games.\footnote{We use the game codes to represent the real-world games due to the business policies.}

\subsubsection{Data}
In the offline experiments, our dataset consists of 8 million users, each labeled to indicate whether they registered the game. The data is partitioned into a training set, a validation set, and a test set, with a 10:1:1 ratio. For the online experiments, we collect data from the 14 days before the advertising system is taken online. We employ the training data from all games to train \textit{a unified model} and subsequently assess the trained model's performance by testing it on each game individually. The related information about dataset is summarized in Table~\ref{tab:data_info}.

\subsubsection{Features}
Following prior work, we use task-specific features, including both dense and sparse features. The dense features include metrics such as the duration of user activity in the game and the amount of money spent. The sparse features encompass basic information like gender and age, as well as interactions with the game over different periods, such as clicks, downloads, registrations. Additionally, features from the item side are also included, such as the game's ID and category, whether it is a strategy, casual, or another type of game.
Additionally, we augment these features by adding our proposed UGL representation as dense features to the task-specific models.

\subsubsection{Task Models} 
In the offline experiments, we enhance the following base models by incorporating our UGL representations:
\vspace{0.2cm}
\begin{itemize}
    \item MLP: All features are directly added to multi-layer perceptrons.
    \item DeepFM: The model creates an end-to-end learning framework emphasizing both low- and high-order feature interactions. It combines the strengths of factorization machines for recommendation and deep learning for feature learning~\cite{guo2017deepfm}.
    \item DIN: The model utilizes a local activation unit to adaptively learn the representation of user interests from historical behaviors concerning a specific item~\cite{zhou2018deep}.
    \item IFM: The model employs Input-aware Factorization Machine, learning a unique input-aware factor for the same feature in different instances via a neural network~\cite{yu2019input}.
    \item DeepLight: The model explores informative feature interactions and prunes redundant parameters at different levels of neural networks~\cite{deng2021deeplight}.
\end{itemize}
\vspace{0.2cm}
In the online experiments, we adopt DIN as our base model.

\subsubsection{Other User Representation Models} 
We compare our proposed UGL representation with other existing user representation models:
\vspace{0.2cm}
\begin{itemize}
    \item PTUM: The model is trained with two self-supervised tasks, namely masked behavior prediction and next $K$ behavior prediction, to enhance the model's training~\cite{wu2020ptum}.
    \item UserBERT: The model is trained with self-supervised tasks, including masked behavior prediction and Behavior Sequence Matching, on unlabeled user behavior data to improve user modeling~\cite{wu2022userbert}.
    \item CLUE: The model employs two Siamese networks to learn user-oriented representations. The optimization goal is to maximize the similarity of learned representations of the same user by two encoders~\cite{cheng2021learning}.
\end{itemize}
\vspace{0.2cm}
Note that to facilitate a rigorous comparison of model-level performance, i.e., the contrastive-learning-based or vanilla-mask-based method with IPM, the input for other user representation methods is kept consistent with our method, where all inputs are assigned the UGL sequence.

\subsubsection{Metrics} 
We employ the Area Under the Curve (AUC) as the offline metric and the Conversion Rate (CVR) as the online metric.

\subsection{Settings for In-Game Item Recommendation}
In-game item recommendation is a classification task that determines whether users will purchase the recommended in-game items. We employ the real-world popular video game, \textit{A}, as our experimental game.

\subsubsection{Data} 
In the offline experiments, as reported in Table~\ref{tab:data_info}, we collect one-day data comprising 0.4 million users with corresponding labels indicating whether they purchased the in-game items or not. The data is partitioned into a training set, a validation set, and a test set with a 10:1:1 ratio. For the online experiments, we gather data one day before the recommendation system is deployed online. 

\subsubsection{Features}
In addition to the features utilized in game advertising, we incorporate additional game-specific features in game \textit{A} to encompass the user in-game, game-specific interests. For instance, game character preferences, character levels, match outcomes, and in-game store expenditures. Besides, features from the item side are also included, such as its ID, attributes, level and price.

\subsubsection{Task Models} 
In both offline and online experiments, we integrate DIN as the base model with our UGL representations.

\subsubsection{Metrics} We use Area Under the Curve (AUC) as the offline metrics and Average Revenue Per Use (ARPU) as the online metrics.

\subsection{Settings for Online Experiments}
We conduct online A/B tests for game advertising and in-game item recommendation from 2023/02/09 to 2023/02/15, a week after the date when the training data is collected. Within the set of users, we divide them into two groups of equal size. The base model is applied to the first group of users, while the model equipped with our UGL representations is applied to the second group of users.

\subsection{Results of Offline Experiments}
Table~\ref{tab:offline_advertising} shows the offline results of game advertising. Models with UGL representation outperform those without UGL representation, showing improvements of 1.53\%-2.08\% in AUC across various games. These increases are consistent in popular, moderate, and cold games. As the base models progress from MLP to DeepLight, the models with our UGL representation demonstrate better performance. We claim that our proposed UGL representation proves to be beneficial for game advertising in model-agnostic settings.

\begin{table*}[t]
\begin{center}
  \resizebox{\columnwidth}{!}{
  \begin{tabular}{lccccccc|c}
    \toprule
    \multirow{2}{*}{Model} & \multicolumn{2}{c}{Popular Games} & \multicolumn{3}{c}{Moderate Games} & \multicolumn{2}{c|}{Cold Games} &  \\
    & \textit{A} & \textit{B} & \textit{C} & \textit{D} & \textit{E} & \textit{F} & \textit{G} & Average\\
    \midrule  	 
    MLP &  0.9170 & 0.7147 & 0.9268 & 0.7568 & 0.8747 & 0.8766 & 0.9096 & 0.8537  \\
    ~~~~~~~~w/ UGL rep. & \textbf{0.9321} & \textbf{0.7354} & \textbf{0.9350} & \textbf{0.8049} & \textbf{0.8951} & \textbf{0.8938} & \textbf{0.9252} & \textbf{0.8745} \\
    \midrule  	 
    DeepFM & 0.9205 & 0.7166 & 0.9308 & 0.7596 & 0.8749 & 0.8839 & 0.9150 & 0.8573 \\
    ~~~~~~~~w/ UGL rep. & \textbf{0.9323} & \textbf{0.7365} & \textbf{0.9582} & \textbf{0.7696} & \textbf{0.8944} & \textbf{0.9041} & \textbf{0.9298} & \textbf{0.8750} \\
    \midrule
    DIN & 0.9351 & 0.7168 & 0.9325 & 0.7669 & 0.8806 & 0.8778 & 0.9162 & 0.8608  \\
    ~~~~~~~~w/ UGL rep. & \textbf{0.9505} & \textbf{0.7249} & \textbf{0.9531} & \textbf{0.7895} & \textbf{0.8907} & \textbf{0.8878} & \textbf{0.9366} & \textbf{0.8761} \\
    \midrule  	 
    IFM  & 0.9255 & 0.718 & 0.9373 & 0.7616 & 0.8780 & 0.8844 & 0.9141 & 0.8598  \\
    ~~~~~~~~w/ UGL rep. & \textbf{0.9474} & \textbf{0.7351} & \textbf{0.9576} & \textbf{0.7783} & \textbf{0.8987} & \textbf{0.9004} & \textbf{0.9269} & \textbf{0.8778} \\
    \midrule  	 
    DeepLight  & 0.9190 & 0.7211 & 0.9313 & 0.7592 & 0.8775 & 0.8790 & 0.9137 & 0.8572  \\
    ~~~~~~~~w/ UGL rep. & \textbf{0.9322} & \textbf{0.7373} & \textbf{0.9490} & \textbf{0.7715} & \textbf{0.8934} & \textbf{0.8983} & \textbf{0.9301} & \textbf{0.8731} \\
    \bottomrule
  \end{tabular}
  }
\caption{Offline results (AUC) of game advertising provided by various models across games. w/ UGL rep. indicates that the base models are equipped with UGL representations as additional features. The best scores are bold.}
\label{tab:offline_advertising}
\end{center}
\end{table*}


Table~\ref{tab:offline_exist} shows the offline results of game advertising for game \textit{A}, \textit{D} and \textit{F} by comparing our proposed UGL representation to other existing user representations. We incorporate various user representations into MLP. Our UGL representations outperform existing methods with the improvements of 1.33-2.26\% in AUC. In comparison to the vanilla mask strategy (PTUM and UserBERT) and contrastive learning (CLUE), our customized method generates high-quality game user representations, effectively addressing the sparsity and imbalance issues inherent in game scenarios.

\begin{table}[t]
  \centering
    \begin{tabular}{lccc}
    \toprule
    Model & \textit{A} & \textit{D} & \textit{F} \\
    \midrule
    MLP w/ PTUM & 0.9238 & 0.7712 & 0.8786  \\
    MLP w/ UserBERT & 0.9251 & 0.7883 & 0.8832   \\
    MLP w/ CLUE &  0.9265 & 0.7698 & 0.8840   \\ 
    \midrule
    MLP w/ UGL Rep. & \textbf{0.9321} & \textbf{0.8049} & \textbf{0.8938}   \\
    \bottomrule
  \end{tabular}
  \caption{Offline results (AUC) of game advertising provided by MLP equipped with our UGL representations or other user representations. The best scores are bold.}
  \label{tab:offline_exist}
\end{table}

The offline results of in-game item recommendation are that DIN equipped with UGL obtain 94.08\% AUC, outperformsn that without UGL representations (93.52\% AUC) by 0.5\% AUC in game \textit{A}. 
The improvement is marginal in the task of in-game item recommendation compared to the task of game advertising. This is attributed to the complexity of the in-game item recommendation task, which involves more intricate user behaviors and potential actions by mistake in the game.

\subsection{Results of Online Experiments}
The first row of the results in Table~\ref{tab:online} shows the online CVR of game advertising. The number of registered users captured by the model with UGL representations is 21.67\% higher than that of the model without it. Specifically, the model with our UGL representations yields average improvements of 18.28\%, 23.40\%, and 22.46\% for popular, moderate, and cold games, respectively. 

The second row of  Table~\ref{tab:online} depicts the online ARPU of in-game item recommendation. The users captured by the model with our UGL representations make 0.82\% more purchases than those without UGL representations.

\begin{table*}[t]
\begin{center}
  \resizebox{\columnwidth}{!}{
  \begin{tabular}{ccccccccc}
    \toprule
    \multirow{2}{*}{} & \multicolumn{2}{c}{Popular Games} & \multicolumn{3}{c}{Moderate Games} & \multicolumn{2}{c}{Cold Games} &  \\
    & \textit{A} & \textit{B} & \textit{C} & \textit{D} & \textit{E} & \textit{F} & \textit{G} & Average\\
    \midrule  	  
     Game advertising & +24.94\%  & +11.61\% & +28.06\% & +6.49\% & +35.64\% & +20.70\% & +24.22\% & +21.67\% \\ 
     \midrule
     In-game item rec. & +0.82\% & / & / & / & / & / & / & / \\
    \bottomrule
  \end{tabular}
  }
\caption{Online improvements (CVR) of game advertising and online improvements (ARPU) of in-game item recommendation provided by DIN with UGL representations across games via A/B tests. The improvements are statistically significant with p $<$ 0.05 under t-test.}
\label{tab:online}
\end{center}
\end{table*}

\section{Analysis}
UGL representations demonstrate the superiority in capturing patterns of user game behaviors that reflect both short and long-term interests in games. To analyze this superiority, we conduct ablation studies by removing various components to assess the impact of our proposed method.

For a comprehensive analysis, we use game \textit{A}, \textit{D} and \textit{F} as representatives of popular, moderate, and cold games, respectively. This allows us to analyze UGL modeling, action augmentation, IPM strategy with different masking rates in training, pooling techniques in inference, and the training of downstream tasks, in the offline experiments of the game advertising task.

\begin{table}[t]
  \centering
    \begin{tabular}{lccc}
    \toprule
    Model & \textit{A} & \textit{D} & \textit{F} \\
    \midrule
    MLP w/ raw seq. &  0.9232 &	0.7678 & 0.8802  \\ 
    MLP w/ random rep. &  0.9105 & 0.7436 & 0.8793   \\ 
    MLP w/ UGL seq. &  0.9265 & 0.7785 & 0.8852   \\ 
    \midrule
    Training w/o IPM & 0.9218 & 0.7596 & 0.8802  \\
    \midrule
    UGL w/o negative feed. & 0.9292 & 0.7898 & 0.8876   \\
    UGL w/o aggregation & 0.9245 & 0.7850 & 0.8830  \\
    \midrule
    Inference w/o max pool. & 0.9255 & 0.7822 & 0.8826  \\
    Inference w/o min pool. &  0.9260 & 0.7898 & 0.8808   \\ 
    Inference w/o avg pool. & 0.9283 & 0.7904 & 0.8852   \\
    Inference w/o var pool. & 0.9290 & 0.7798 & 0.8870   \\
    \midrule
    Ours & \textbf{0.9321} & \textbf{0.8049} & \textbf{0.8938}   \\
    \bottomrule
  \end{tabular}
  \caption{Results (AUC) of the ablation studies based on MLP-based variants. Ours is the MLP model with UGL representations, IPM training and inference with the complete pooling strategies. The best scores are bold.}
  \label{tab:ablation}
\end{table}

\subsection{Large-Scale UGL modeling}
In our experiments, the proposed UGL representations are added to base models as extra dense features. The number of parameters in the base models is fewer than the models with UGL representations. To demonstrate that the increased performance comes from the UGL representation instead of the added parameters, we compare the base model with UGL representations (Ours) to the model equipped with raw action sequences of users (w/ raw seq.) and the model with random representations of users (w/ random rep.). All the models have the same size of parameters.

As shown in the first group of Table~\ref{tab:ablation}, 
the model equipped with the raw sequence as input features instead of the UGL sequence has an average AUC decrease of 1.98\%. This indicates that the raw sequence fails to provide the information of the user behavior history. 
Also, when the model employs a random representation sampled from a standard normal distribution, the AUC for the three games drops by an average of 3.24\%. This demonstrates that the improvements stem from the UGL rep. rather than the increased parameter size of the model. 
Furthermore, when replacing the raw sequence with UGL sequence (w/ UGL seq.), the variant gains an improvement of 0.77\%, which further indicates the constructed UGL can provide more effective information than the raw sequence.


Large-scale training on our model is substantial. A larger training dataset enables our model to learn more intricate patterns and nuances in user behavior, resulting in improved performance. This is particularly crucial in the context of game-related applications, where user actions and preferences can be diverse and dynamic. The scale of our training data contributes significantly to the model's ability to generalize and capture a broader spectrum of user interactions with games.

\subsection{Inverse Probability Masking} 

As shown in the second group of Table~\ref{tab:ablation}, the model without IPM experiences an average AUC decline of 2.75\%. It shows the critical role of the IPM strategy in enhancing the quality of learned representations. Training with IPM, our model becomes more adept at capturing intricate patterns in user behavior, particularly in scenarios with game imbalance, ultimately leading to improved performance.

\begin{figure}[t]
\begin{center}
\includegraphics[width=0.8\columnwidth]{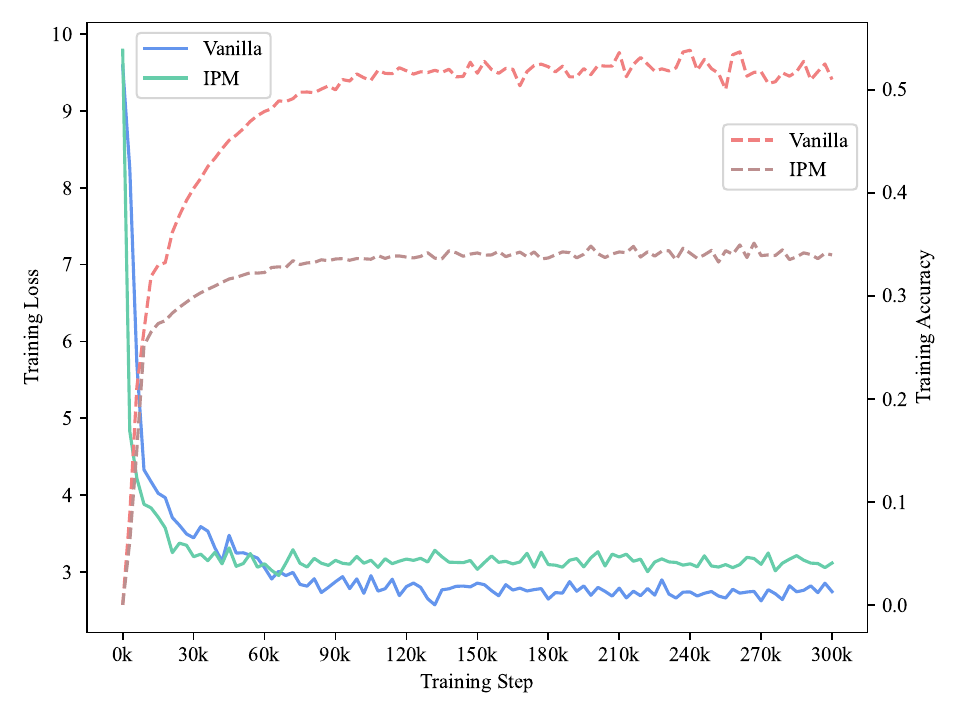}
\caption{Training losses (solid) and accuracy (dashed) of UGL representation with IPM and with vanilla methods.}
\label{fig::bertloss}
\end{center}
\end{figure}

As shown in Figure~\ref{fig::bertloss}, the self-supervised task with the vanilla masking strategy exhibits lower loss and higher accuracy in representation learning compared to the task with the Inverse Probability Masking (IPM) strategy, indicating that the self-supervised task with IPM is more challenging. The self-supervised task with vanilla masking strategy is relatively easy, potentially causing the model to struggle to learn valuable representations of user behaviors effectively. However, the IPM strategy is applied during training to selectively mask out certain actions in the user sequences, creating a scenario where the model needs to infer the missing information. This process encourages the model to \textit{be fair} to focus on the relevant context and dependencies within the sequences, leading to more meaningful and contextually rich representations.

\subsection{Action Augmentation}
The action augmentations are key to constructing UGL sequences. The negative feedback strategy is applied to generate more actions for a UGL sequence, effectively doubling its length, while the aggregation is applied to condense repeated consecutive actions, resulting in a shorter sequence. To investigate the contribution of these two augmentation strategies, we remove these components and retrain the UGL representations added to the base models.

As shown in the third group of Table~\ref{tab:ablation}, when user representation learning is conducted on the sequence without Aggregation (w/o aggregation) or Negative Feedback (w/o negative feed.), the final performance declines by an average of 1.27\% or 0.81\%, respectively. This indicates that both strategies are beneficial for capturing user interests more effectively. Notably, the aggregation strategy, which shortens the sequence, contributes more significantly to the UGL sequence modeling.

Action augmentations are noteworthy in enhancing the robustness and generalization capabilities of our model. This augmentation aids in mitigating overfitting and capturing a more comprehensive range of patterns in user interactions with games.
Incorporating action augmentation contributes to the ability of models to handle variations in user behaviors, especially in dynamic and unpredictable gaming scenarios. The improved generalization achieved through action augmentation is crucial for the model to perform effectively in real-world applications where user actions may exhibit considerable diversity and complexity.

\subsection{Mask Rates}\label{subsec::diff_mask_rate} 
UGL representations are trained using a self-supervised task where tokens are masked with probabilities controlled by the coefficients $\alpha$, along with constant probabilities $q^v$ and $q^c$. Here, $q^v$ represents the masking probability in the vanilla masking strategy, and $q^c$ represents the bounding probability in the IPM strategy. To investigate how the hyperparameters $q^v$ and $q^c$ affect the performance of UGL representation learning, we assign them different values and analyze the model's performance on downstream tasks.

\begin{figure}[t]
\begin{center}
\includegraphics[width=0.8\columnwidth]{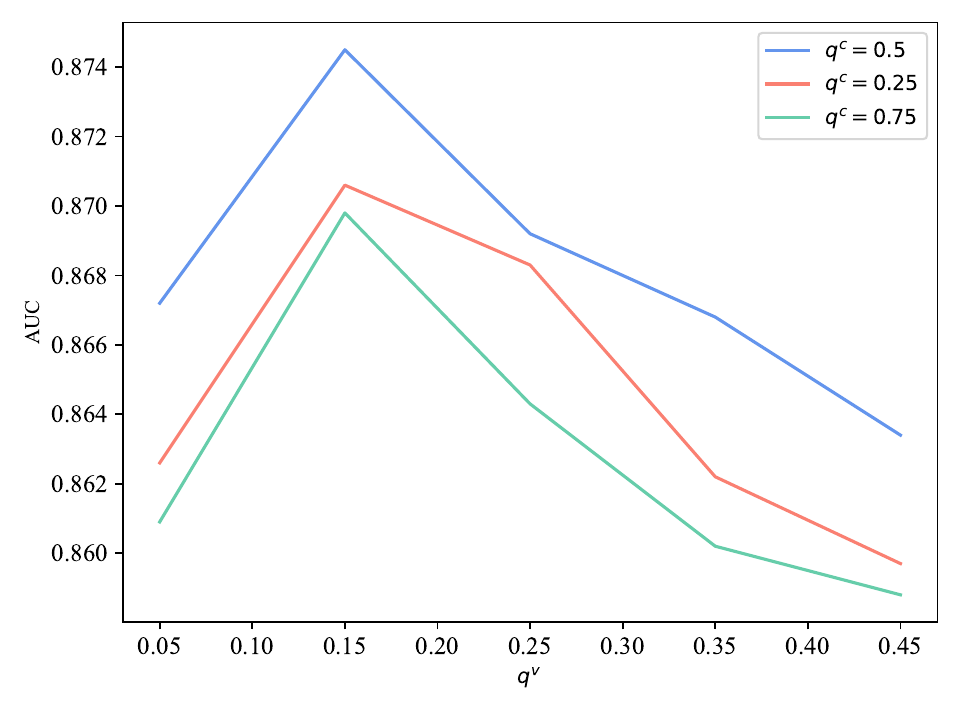}
\caption{Results (AUC) of the game advertising with the different $q^v$ and $q^c$.
}
\label{fig::maskrate}
\end{center}
\end{figure}

As shown in Figure \ref{fig::maskrate}, both excessively low and high values for $q^v$ can lead to a decline in performance. 
In low-masking scenarios, excessive context preservation creates a trivial prediction task that fails to sufficiently stimulate the model's pattern discovery capabilities, resulting in compromised learning efficacy.
Conversely, in high-masking conditions, the severe fragmentation of behavioral context triggers an information bottleneck effect, where the sparsity of observable signals renders the masked action recovery task challenging.
Therefore, selecting an appropriate mask rate is crucial to maintains an equilibrium between contextual signal preservation and prediction challenge complexity.


\subsection{Pooling Strategies} \label{subsec::diff_pooling_tech}

Pooling strategies serve as a critical architectural component in sequential representation learning, primarily designed to extract and aggregate discriminative features from temporal data structures. We hypothesize that heterogeneous pooling mechanisms encode complementary information patterns through their distinct temporal aggregation properties. To empirically validate the functional importance of these design choices, we conduct systematic ablation studies evaluating the performance degradation caused by removing each pooling components, thereby quantifying their individual contributions to model efficacy.

As quantitatively demonstrated in the fourth group of Table~\ref{tab:ablation}, the absence of any pooling strategy induces statistically significant performance degradation, manifesting as 1.06-1.59\% reductions in average AUC metrics. These findings empirically validate the criticality of multi-strategy pooling mechanisms in preserving task-relevant information. The implemented pooling hierarchy facilitates hierarchical extraction of multi-granular behavioral signatures from user interaction sequences, enhancing its ability to discern essential patterns in user behavior.


\subsection{Training of Task-specific Models}
The task-specific models are trained with the training data for respective tasks. The various base models exhibit differing capabilities primarily due to their adoption of distinct methods to extract features from the input data. These extracted features then interact with each other, forming high-order feature representations through various methods. To investigate whether the feature interactions in the base models encompass the UGL representations, we visualize the training loss in different base models.

\begin{figure*}[htbp]
  \centering
  \subfigure[MLP]{
    \includegraphics[width=0.25\textwidth]{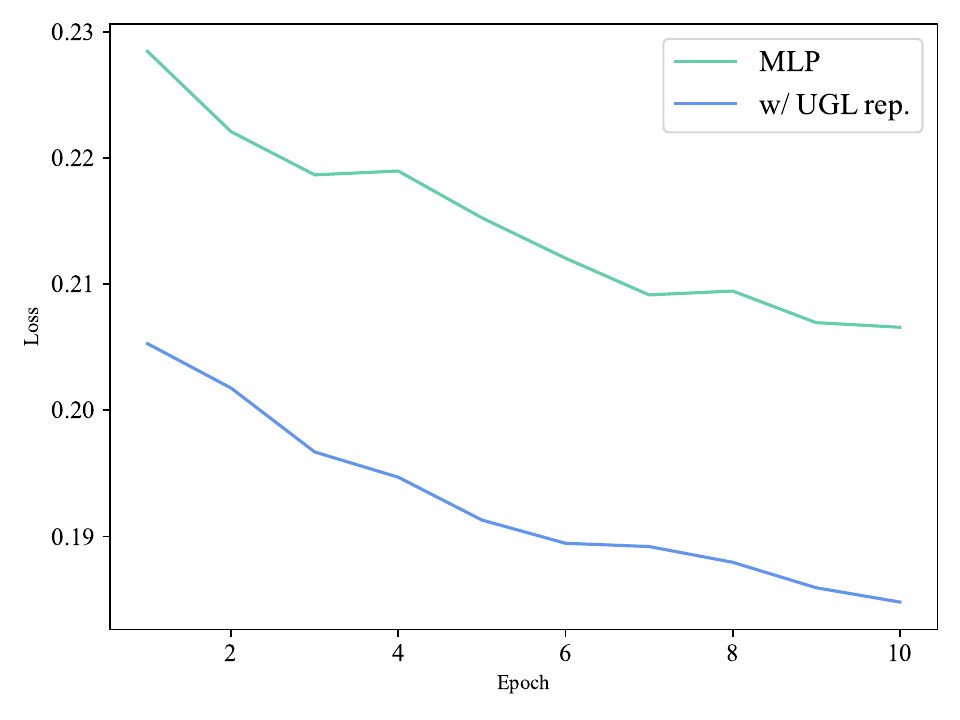}
  }
  \subfigure[DeepFM]{
    \includegraphics[width=0.25\textwidth]{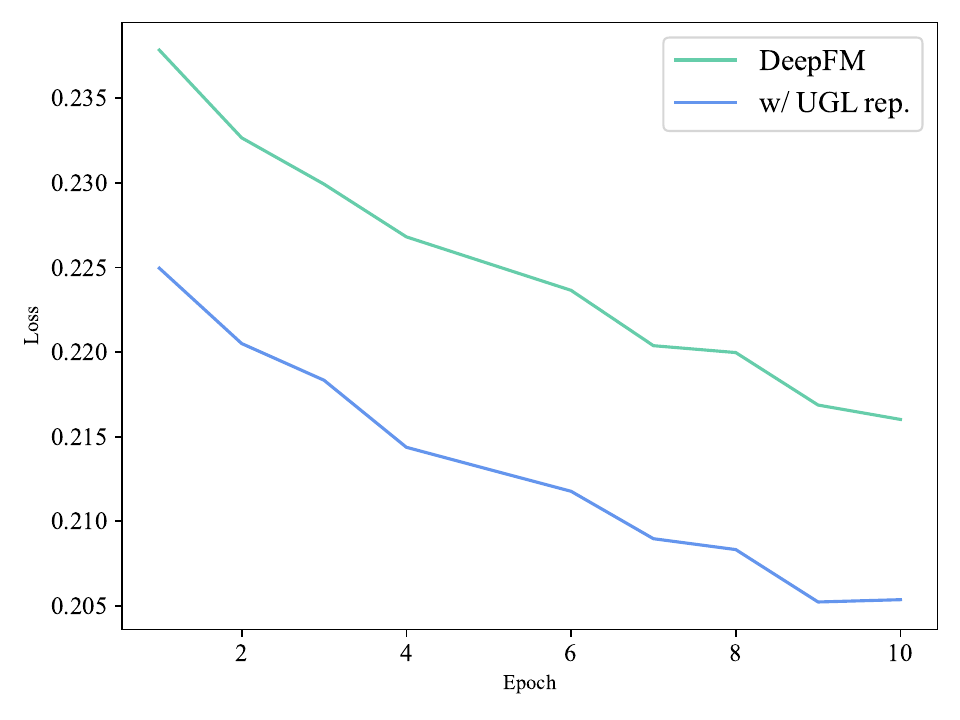}
  }
  \subfigure[DIN]{
    \includegraphics[width=0.25\textwidth]{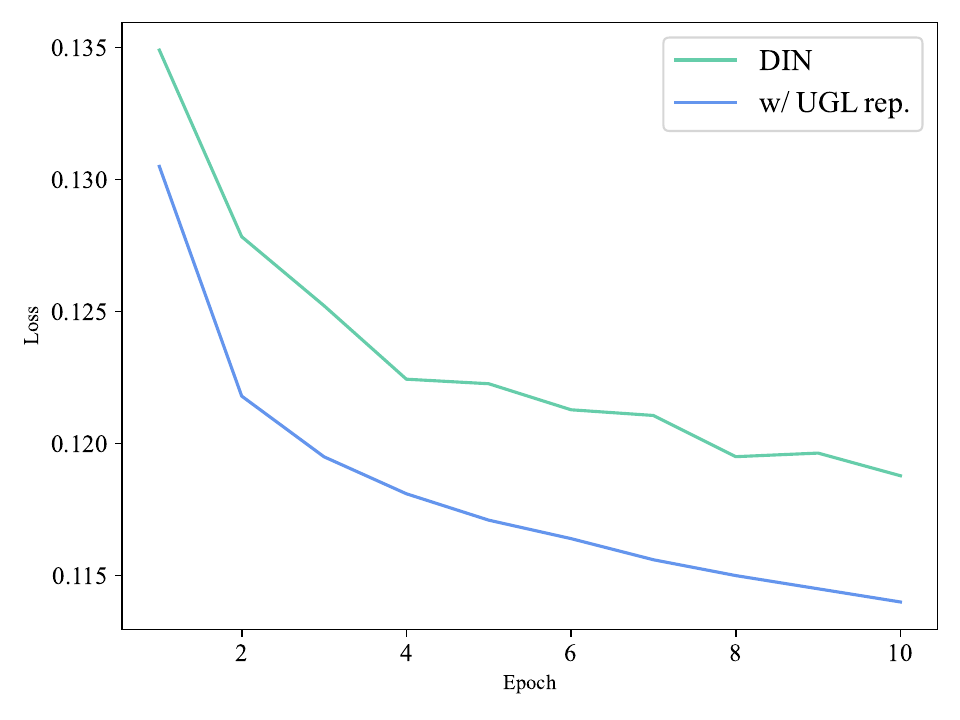}
  }
  \subfigure[IFM]{
    \includegraphics[width=0.25\textwidth]{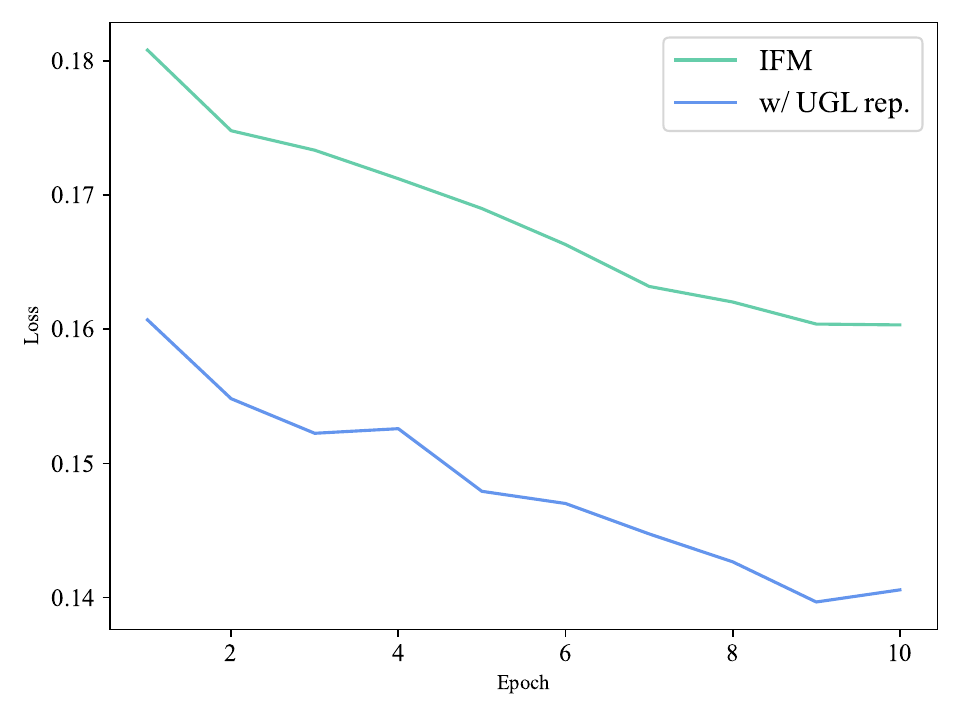}
  }
  \subfigure[DeepLight]{
    \includegraphics[width=0.25\textwidth]{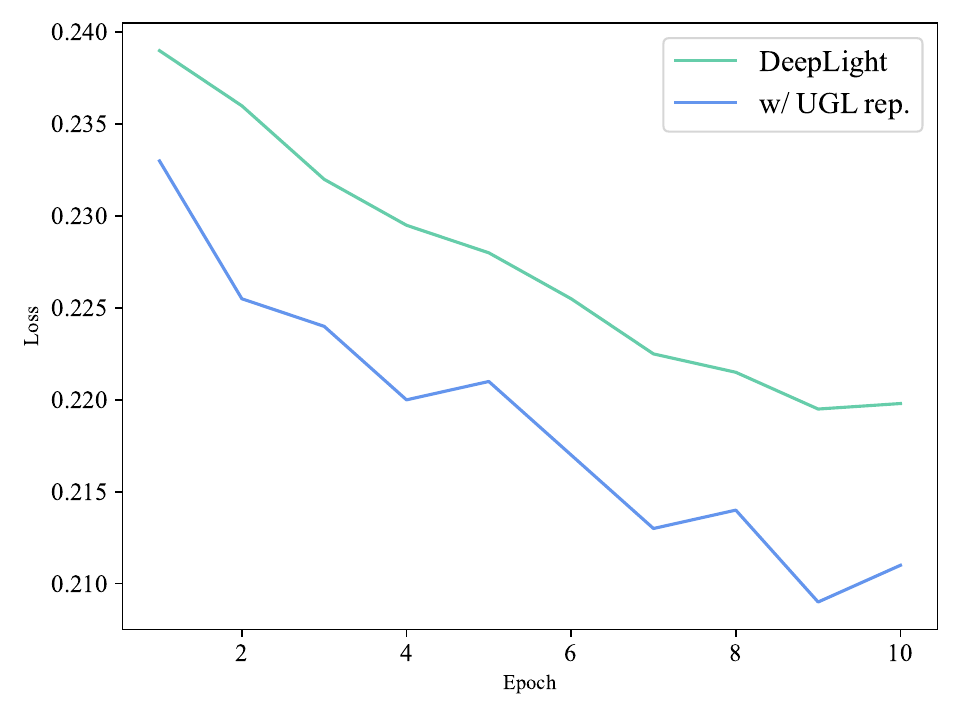}
  }
  \caption{Training loss of the models for game advertising with or without UGL representations.}
  \label{fig:train_loss}
\end{figure*}

As illustrated in Figure~\ref{fig:train_loss}, the convergence curves demonstrate a consistent performance gap between baseline implementations and their UGL-enhanced counterparts across all architectural variants in game advertising scenarios. The significant reduction in training loss achieved by UGL-enhanced models substantiates that our proposed representations provide substantial complementary information orthogonal to existing feature interaction paradigms. This empirical evidence suggests that UGL-derived features effectively augment the representational capacity of diverse model architectures through synergistic combination with their intrinsic feature interaction mechanisms.


\section{Conclusion}
In this work, we proposed the User Game Lifecycle (UGL) along with action augmentation strategies, namely, aggregation and negative feedback. These strategies diversify and enrich user behaviors, providing a more comprehensive representation of their interests, aiming to address the challenges posed by game sparsity.
Additionally, we introduced the Inverse Probability Masking (IPM) strategy to facilitate a balanced representation optimization process in both popular and cold games. This strategy aims to mitigate the challenges of game imbalance in game user representation learning.
Both real-world offline and online experiments were conducted on game advertising and in-game recommendation tasks. The experimental results demonstrate that the proposed UGL representations effectively capture user interests across games, achieving significant improvements compared to existing methods of user representations.

\bibliographystyle{elsarticle-num}
\bibliography{sample-base}




\end{document}